\pdfoutput=1

\documentclass[11pt]{article}

\usepackage[]{acl}

\usepackage{times}
\usepackage{latexsym}
\usepackage{amssymb}
\usepackage[T1]{fontenc}
\usepackage[utf8]{inputenc}

\usepackage{microtype}

\usepackage{graphicx}
\usepackage{multirow}
\usepackage{adjustbox}
\usepackage{booktabs}
\usepackage{array}
\usepackage{caption}
\usepackage{subcaption}
\usepackage{svg}
\usepackage{forest}
\usepackage{enumitem}
\usepackage{rotating}

\usepackage{arydshln}

\usepackage{CJKutf8}

\usepackage{xcolor,colortbl}
\usepackage{makecell} 

\usepackage{soul}
\definecolor{hlgreen}{HTML}{B2D5CB}
\definecolor{hlblue}{HTML}{ADD8E6}
\definecolor{hlyellow}{HTML}{EADDCA}

\definecolor{color1}{HTML}{D6E7D0}
\definecolor{color2}{HTML}{CFE2F3}
\definecolor{color3}{HTML}{F4CCCC}
\definecolor{color4}{HTML}{FDF1CB}

\usepackage{xspace}

\DeclareSymbolFont{extraup}{U}{zavm}{m}{n}
\DeclareMathSymbol{\varheart}{\mathalpha}{extraup}{86}

\title{Advancing Singlish Understanding: \\ 
Bridging the Gap with Datasets and Multimodal Models
}

\author{
\textbf{
Bin Wang*\textsuperscript{$\diamondsuit$} 
\ \quad 
Xunlong Zou*\textsuperscript{$\diamondsuit$} 
\ \quad 
Shuo Sun\textsuperscript{$\diamondsuit$} 
\ \quad
Wenyu Zhang\textsuperscript{$\diamondsuit$} 
\ \quad 
Yingxu He\textsuperscript{$\diamondsuit$} 
}
\\
\textbf{
Zhuohan Liu\textsuperscript{$\diamondsuit$} 
\ \quad 
Chengwei Wei\textsuperscript{$\diamondsuit$} 
\ \quad 
Nancy F. Chen\textsuperscript{$\diamondsuit,\dag$}
\ \quad 
AiTi Aw\textsuperscript{$\diamondsuit$} 
} \\
\textsuperscript{$\diamondsuit$}Institute for Infocomm Research (I$^2$R), A*STAR, Singapore\\
\textsuperscript{$\dag$}Centre for Frontier AI Research (CFAR), A*STAR\\
\texttt{wang\_bin@i2r.a-star.edu.sg} \\
}

\begin{document}
\maketitle

\begin{abstract}

    Singlish, a Creole language rooted in English, is a key focus in linguistic research within multilingual and multicultural contexts. However, its spoken form remains underexplored, limiting insights into its linguistic structure and applications. To address this gap, we standardize and annotate the largest spoken Singlish corpus, introducing the \textbf{Multitask National Speech Corpus (MNSC)}. These datasets supports diverse tasks, including Automatic Speech Recognition (ASR), Spoken Question Answering (SQA), Spoken Dialogue Summarization (SDS), and Paralinguistic Question Answering (PQA). We release standardized splits and a human-verified test set to facilitate further research. Additionally, we propose \textbf{SingAudioLLM}, a multi-task multimodal model leveraging multimodal large language models to handle these tasks concurrently. Experiments reveal our models’s adaptability to Singlish context, achieving state-of-the-art performance and outperforming prior models by 10–30\% in comparison with other AudioLLMs and cascaded solutions.\footnote{
    To facilitate future work, we released our datasets, model, and code at \url{https://github.com/AudioLLMs/Singlish} 
    \\
    $^*$Equal Contributions
    }

\end{abstract}

\section{Introduction}

    Speech technologies have evolved over decades, progressing from modularized solutions for speech recognition~\cite{povey2011kaldi,radford2023robust}, speaker identification~\cite{togneri2011overview}, and gender recognition~\cite{hechmi2021voxceleb} with modularized toolkits like Kaldi~\cite{povey2011kaldi} and ESPnet~\cite{watanabe2018espnet} to advanced solutions integrating large language models for multimodal understanding in an all-encompassing, omni-style approach~\cite{team2023gemini,zhang2024mowe,li2024baichuan}. Nevertheless, the combination of speech and language technology is deemed the trend for solving more complex multimodal tasks requiring a holistic understanding of information from different sources.

    In the field of speech and audio understanding, there is a growing trend and initial success of utilizing large language models for complex and direct reasoning over audio signals, which we refer to as Audio Large Language Models (AudioLLMs)~\cite{cui2024recent,chu2024qwen2,nguyen2024spirit,he2024meralion}. This approach offers several advantages over traditional modular solutions~\cite{wang2024audiobench}, including but not limited to: 1) the ability to leverage pre-trained models for more sophisticated reasoning, 2) faster decoding time by avoiding multiple decoding steps, as in end-to-end models, 3) reduced error propagation between modules, and 4) the capacity for a more holistic understanding of audio signals. Building on this paradigm, our work explores the application of AudioLLMs for Singlish spoken understanding~\cite{deterding2007singapore}, aiming to address the unique characteristics of Singlish, which include its multilingual nature, diverse accents, and complex syntactic structures.

    \begin{figure*}[t]
         \centering
             \includegraphics[width=0.93\textwidth]{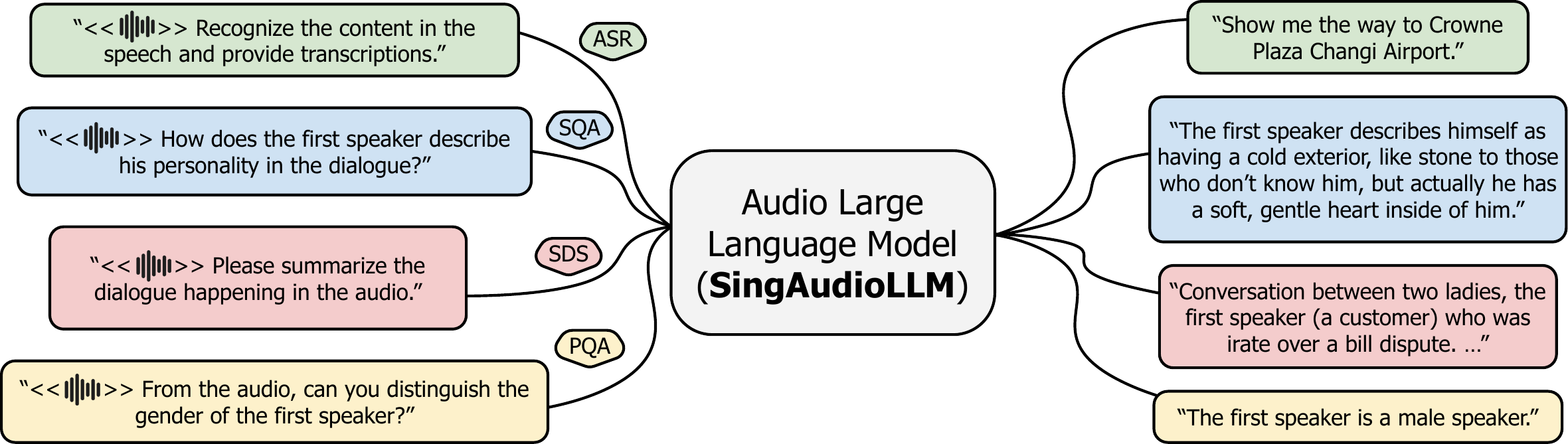}
            \caption{A demonstration of our target multi-task learning datasets and end-to-end multimodal models for understanding spoken Singlish. The tasks include Automatic Speech Recognition (ASR), Spoken Question Answering (SQA), Spoken Dialogue Summarization (SDS), and Paralinguistic Question Answering (PQA).}
            \label{fig:overview_singaudio}
    \end{figure*}

    Existing Singlish spoken corpora have primarily focused on linguistic analysis and speech recognition tasks~\cite{deterding2001nie,chen2010development,lyu2010seame,tanspontaneous}. Given the relatively small population of Singlish speakers, estimated at just a few million, resources for Singlish speech corpora are significantly more limited compared to major languages like English, Chinese, French, and Spanish. Singapore’s government agency, IMDA, has open-sourced the largest available Singlish corpus, known as the National Speech Corpus~\cite{koh2019building}. Despite its great value, this corpus still has several limitations: inconsistent recording scenarios across its components, the absence of standardized splits for benchmarking, and a sole focus on speech recognition tasks. To address these issues, we establish standardized multitask spoken understanding datasets, expanding beyond basic ASR to include sentence and dialogue speech recognition, spoken question answering, spoken dialogue summarization, and paralinguistic question answering. These datasets are developed through metadata extraction, cleaning of the existing corpus, and synthesis using large language models. All test sets are either pre-annotated or post-edited by our human annotators to ensure high quality and reliability. The datasets, collectively named the \textbf{Multitask National Speech Corpus (MNSC)}, are publicly released to support future research on Singlish spoken understanding tasks. This resource is especially valuable in the era of multimodal large language models, which aim to handle multiple tasks within a single unified framework. An illustration can be found in Figure~\ref{fig:overview_singaudio}.

    To enhance analysis and set a robust baseline for future benchmarks, we investigate best practices for leveraging AudioLLMs with different architectural designs. Our model, SingAudioLLM, utilizes joint training across multiple tasks. The results demonstrate that the fusion model achieves state-of-the-art (SOTA) performance, outperforming both generalized AudioLLMs and specialized speech recognition models (e.g., Whisper) by over 10\%. Additionally, we analyzed the impact of audio encoder selection, adaptors, and pre-trained LLMs, offering insights into which components contribute most significantly to performance across various tasks.

    \noindent Our major contributions are as follows:
    \begin{itemize}[noitemsep, topsep=0pt, left=0pt]
        \item We present the \textbf{MNSC} datasets for Singlish spoken understanding, featuring four tasks with standardized training datasets and human-annotated test sets. Baselines are established across all tasks to support future benchmarking and research advancements. MNSC marks the largest well-organized resource for Singlish-specific spoken language processing, enabling comprehensive evaluation and fostering innovation in multilingual and code-switched NLP research.

        \item We developed \textbf{SingAudioLLM}, a multimodal large language model designed for diverse Singlish spoken understanding tasks. The model incorporates joint training across multiple tasks, leveraging a fusion architecture to maximize performance. Our experiments highlight its ability to capture multilingual nuances, achieving state-of-the-art results across various benchmarks and outperforming prior models by significant margins.
        
        \item We analyzed the impact of various components on the performance of Singlish spoken understanding. Our study evaluated factors such as audio encoders, adaptors, and pre-trained language models to determine their contribution to overall effectiveness. The findings provide insights into which elements play a more critical role in enhancing task-specific performance.        
    \end{itemize}

\section{Related Work}

    \textbf{Audio Large Language Model}. There is a growing trend of integrating large language models with audio understanding capabilities (including speech, music, and more) to handle human instructions conditioned on the audio modality~\cite{peng2024survey,ji2024wavchat,cui2024recent}. Most approaches treat LLMs as a foundation, incorporating audio representations in either continuous or discrete forms~\cite{rubenstein2023audiopalm}. Using discrete tokenizers offers potential advantages, such as extracting more abstract features, reducing the risk of overfitting, and facilitating integration into audio generation scenarios~\cite{defossez2024moshi}. To seamlessly integrate multiple modalities while preserving the LLM's capabilities, cross-modality instruction tuning is applied to enhance the fusion of different modalities effectively or interleaved modality features~\cite{pan2023cosmic,kimparalinguistics,nguyen2024spirit}.

    Fusion models offer several key advantages over cascaded models, including minimizing information loss, avoiding cumulative errors, and enabling faster decoding times. They also hold the potential to handle more complex tasks, such as decoding speaker-specific information and performing context-dependent question answering that incorporates paralinguistic features (e.g., accent, emotion, and gender) or environmental cues. However, existing research efforts are predominantly focused on high-resource languages such as English~\cite{tang2023salmonn,lu2024developing} and Chinese~\cite{chu2024qwen2,yang2024building}. 
    Research on AudioLLMs for low-resource and Creole languages remains limited, largely because these technologies are still in their early stages of development and suitable resources are scarce.

    \begin{table*}[t] 
    \centering
    \begin{adjustbox}{width=0.93\textwidth, center} 
    \begin{tabular}{ | c | c | c | c | c | c | } 
    \toprule 
    \textbf{Category} & \textbf{Source} & \textbf{Split} & \textbf{\#Samples} & \textbf{Total hours} & \textbf{Avg./Min./Max. Len} \\ 
    \hline
    
    \rowcolor{color1!30} \multicolumn{2}{|l|}{\textbf{ASR - Automatic Speech Recognition}} \\ 
    \hline
    
    \multirow{2}{*}{Sentence-Level} & \multirow{2}{*}{NSC-PART 1–2} & Train & 2.3M, 2.5M & 3.3K, 3.2K & 4.9s, 1.2s, 27.4s \\
     &  & Test  & 3K, 3K & 4.0K, 4.9K & 5.4s, 1.9s, 14.9s \\
    \hdashline
    
    \multirow{2}{*}{Dialogue-Level} & \multirow{2}{*}{NSC-PART 3–6} & Train & 96K, 9K, 24K, 104K & 741, 70, 168, 697 & 25s, 15s, 30s \\
     &  & Test & 1K, 1K, 1K, 1K & 7.7, 7.3, 6.9, 6.8 & 25s, 15s, 30s \\
    \hline
    
    \rowcolor{color2!30} \multicolumn{2}{|l|}{\textbf{SQA - Spoken Question Answering}} \\ 
    \hline
    \multirow{2}{*}{Dialogue-Level} & \multirow{2}{*}{NSC-PART 3–6} & Train & 96K, 9K, 24K, 104K & 741, 70, 168, 697 & 25s, 15s, 30s \\
     &  & Test & 100, 100, 100, 100 & 0.8, 0.8, 0.8, 0.8 & 27s, 15s, 30s \\
    \hline
    
    \rowcolor{color3!30} \multicolumn{2}{|l|}{\textbf{SDS - Spoken Dialogue Summarization}} \\ 
    \hline
    \multirow{2}{*}{Dialogue-Level} & \multirow{2}{*}{NSC-PART 3–6} & Train & 96K, 9K, 24K, 104K & 741, 70, 168, 697 & 25s, 15s, 30s \\
     &  & Test & 100, 100, 100, 100 & 0.8, 0.8, 0.8, 0.8 & 27s, 15s, 30s \\
    \hline
    
    \rowcolor{color4!30} \multicolumn{2}{|l|}{\textbf{PQA - Pralinguistic Question Answering}} \\ 
    \hline
    \multirow{1}{*}{Sentence-Level} & \multirow{2}{*}{NSC-PART 1-2} & Train & 4.7M & 6.5K & 4.9s, 1.2s, 27s  \\
    (Gender/Accent)  &  & Test & 6K & 9 & 5.4s, 1.9s, 14.9s \\
   
    \hdashline
    \multirow{1}{*}{Dialogue-Level} & \multirow{2}{*}{NSC-PART 3–5} & Train & 130K & 979 & 27s, 15s, 30s \\
    (Gender/Accent)  &  & Test & 3K & 21.9 & 26.3s, 15s, 30s \\

    \bottomrule 
    \end{tabular} 
    \end{adjustbox} 
    \caption{Statistics for MNSC - Multitask National Speech Corpus.} 
    \label{dataset:mnsc_statistics} 
    \end{table*} 

    \textbf{Singlish Spoken Understanding}. Previous research on Singlish spoken corpora has primarily focused on ASR~\cite{deterding2001nie,lyu2010seame,boo2023particle}, code-switching tasks~\cite{lyu2010seame,koh2019building}, and linguistic pattern analysis~\cite{lim2011tone,goh2016anatomy,lee2022analysis}. However, the speech corpus is still limited and contains just a few hundred hours of speech~\cite{chen2010development}. To address this, the Info-communications and Media Development Authority (IMDA), a government agency in Singapore, launched a major initiative called National Speech Corpus (NSC) to collect the first large-scale Singapore English corpus for ASR tasks. Released in multiple stages, the corpus comprises approximately 10,000 hours of speech, predominantly featuring the Singaporean accent (Singlish) and code-switching patterns. Despite the availability of abundant resources under a publicly accessible license, Singlish spoken understanding tasks remain under-explored due to several challenges: 1) the focus is largely limited to automatic speech recognition, 2) the absence of standardized train/test splits for consistent benchmarking, and 3) difficulties in processing data with misaligned annotations. Therefore, in this work, we release the Multitask National Speech Corpus (MNSC), designed to support complex reasoning tasks beyond speech recognition for research and practical applications.

\section{Multitask National Speech Corpus}
\label{sec:4}

    A major challenge in spoken Singlish understanding is the lack of standardized data and limited resources. While the National Speech Corpus~\cite{koh2019building} focuses on ASR, it lacks support and evaluation sets for broader tasks. To address this, we introduce the Multitask National Speech Corpus (\textbf{MNSC}), which extends coverage to dialogue summarization, spoken understanding, and paralinguistic analysis, accompanied by newly created human-labeled test sets and standardized train-test splits. All datasets are publicly available, offering a robust, multitask-oriented resource to advance research in spoken Singlish understanding.
    
    \subsection{Source}

        National Speech Coprus is the first large-scale Singapore English corpus spearheaded by the IMDA from Singapore\footnote{\url{https://www.imda.gov.sg/how-we-can-help/national-speech-corpus}}. Its release is divided into two phases and 9 parts, encompassing approximately 10,000 hours of recorded audio from around 1,000 local Singaporeans. The six parts are organized into distinct themes and conditions, detailed as follows:
        
        \begin{itemize}[noitemsep, topsep=0pt, left=0pt]

            \item \textbf{PART 1}. This part contains prompted recordings of phonetically-balanced scripts, featuring \textbf{standard English sentences} with local accents that reflect the unique linguistic characteristics of the speakers.
            
            \item \textbf{PART 2}. This part includes prompted recordings of sentences randomly generated from topics such as people, food, locations, and brands. The recordings incorporate a rich use of \textbf{local terms and locations} and are accompanied by orthographic transcriptions, highlighting the linguistic and cultural diversity of the context.
            
            \item \textbf{PART 3}. This part features conversational data where speakers engage in discussions covering topics from daily life and structured interactions such as gameplay.
            
            \item \textbf{PART 4}. This part focuses on \textbf{code-switching}, where speakers alternate between Singapore English and their mother tongue languages including Mandarin, Malay, or Tamil.
            
            \item \textbf{PART 5}. This part includes recordings of speakers adopting specific speaking styles, such as debates, finance-related discussions, and expressions of positive and negative emotions.
            
            \item \textbf{PART 6}. This part comprises recordings of speakers participating in scenarios designed around three themes: holiday / hotel / restaurant, bank / telephone / insurance, and HDB / MOE / MSF.
            
        \end{itemize}

        We classify the recordings into monologue and dialogue categories. For dialogue scenarios, audio channels from different rooms or speakers are merged into a single unified channel covering the entire conversation, with a maximum dialogue duration capped at 30 seconds. Recordings that could not be perfectly aligned were excluded from the MNSC dataset.

    \begin{figure*}[t]
         \centering
             \includegraphics[width=0.95\textwidth]{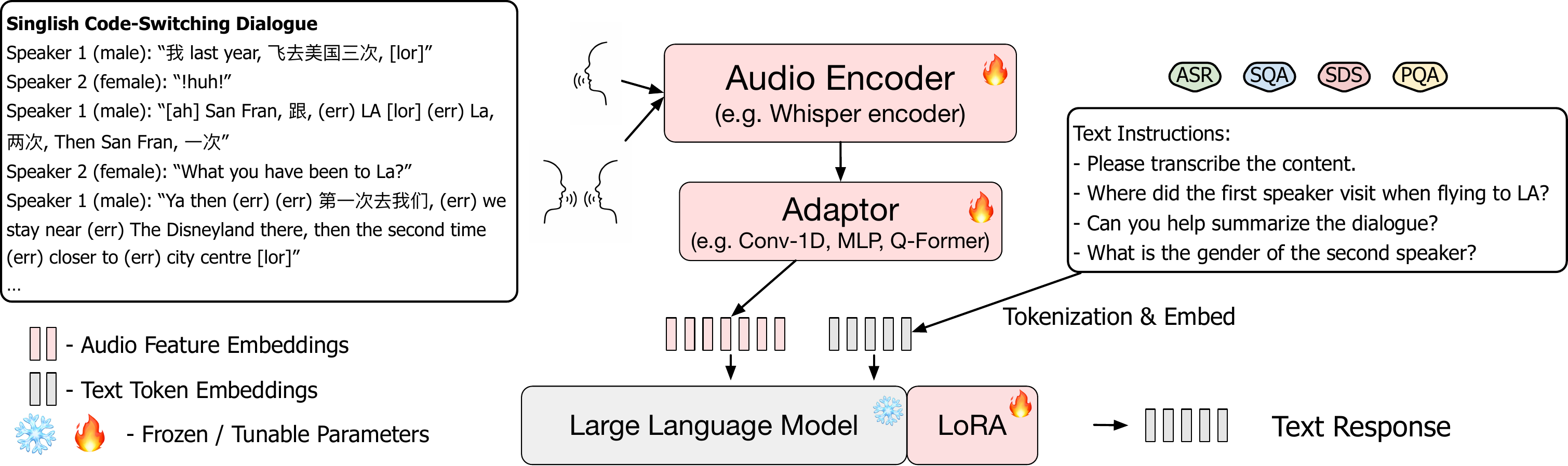}
            \caption{The architecture of SingAudioLLM incorporates a modality adaptor that processes audio features extracted for representation. These adapted features are then combined with textual instructions and input into the large language model, enabling it to handle a variety of tasks effectively.}
            \label{fig:singaudiollm_framework}
    \end{figure*}

    \subsection{Data Generation and Human Annotation}

        For multitasking, we construct the dataset to support four tasks. The complete dataset statistics for each task are presented in Table~\ref{dataset:mnsc_statistics}. For ASR, the transcriptions are split into sentence-level and dialogue-level categories, with dialogue-level transcriptions created by combining original orthographic transcripts.

        For the SQA (Spoken Question Answering) and SDS (Spoken Dialogue Summarization) tasks, we synthesized question-answer pairs and dialogue summaries for training, while employing human annotators to create a high-quality, human-labeled test set. To generate the training data, we utilized the LLM (LLAMA-3.1-70B-Instruct model~\cite{dubey2024llama}), prompting it with transcriptions to elicit meaningful instructions~\cite{wang-etal-2023-self-instruct,wang-etal-2023-instructive}. This form of data augmentation is a commonly adopted practice, enabling the efficient generation of diverse and contextually rich training samples.

        For evaluation, human annotators, who were fully informed about the task and compensated appropriately, conducted detailed annotations, employing cross-validation techniques to ensure quality and consistency. This process yielded 100 test samples for each subtask, culminating in a total of 800 high-quality samples spanning the QA and summarization tasks. In part 4, where code-switching occurs, all data is labeled in English but requires an understanding of multilingual exchanges. This approach ensures annotation consistency while addressing the content's multilingual nature.
               
        For PQA (Paralinguistic Question Answering), we utilize metadata such as speaker ethnicity and gender. This metadata allows us to assess the model's ability to infer beyond basic transcriptions by identifying attributes such as gender and accents, thereby enriching the multitasking framework with paralinguistic understanding. The task is divided into sentence-level and dialogue-level categories, with the former involving a single speaker and the latter encompassing interactions between multiple speakers.

\section{SingAudioLLM}

    As shown in Figure~\ref{fig:singaudiollm_framework}, we explore the fusion model by extracting audio features and adapting the feature to the token embedding space of LLMs through adaptor modules. Such features are concatenated with token embedding of texture instructions and learned jointly with corresponding responses end-to-end. In this section, we will introduce the components leveraged and the training scheme of our AudioLLM for Singlish spoken understanding.

    \subsection{Encoder}

        To understand speech signals, we need to extract features and encode their semantic and acoustic meanings into a representation space. In this context, we utilize the Whisper model, specifically Whisper-Large-v3~\cite{radford2023robust}, due to its robust performance in multilingual automatic speech recognition and speech translation. Given that Singlish is an evolving language incorporating accents and terms from its diverse multilingual cultural mix, leveraging a multilingual encoder as a foundational feature extractor can be advantageous.

        Formally, with a sequence of audio signals as $\mathbf{x} = \{x_{1}, x_{2}, \dots, x_{T}\}$, the audio signal $\mathbf{x}$ is encoded with the encoder of Whisper model $\phi$, which consists of extracting mel-spectrum features and multiple layers of self-attention and feed-forward networks, to hidden representations $\mathbf{h}$.
        $$\mathbf{h}=\phi(\mathbf{x}), \mathbf{h} \in \mathbb{R}^{\tau \times d} $$
        where $d=1280$ is the size of the hidden dimension and $\tau=1500$ is the encoded sequence length. Note for Whisper encoder, the input audio signal is always padded to 30 seconds and the sequence length $\tau$ is fixed and $\tau \ll T$.

    \subsection{Modality Adapter}
    \label{sec:adaptor}

        After encoding, we further adapt the feature through adaptors to match the representation of token embeddings from LLMs. Here, we experiment with three variances: Rescale-MLP, 1-Dimensional Convolutional Layer (Conv-1D)~\cite{li2023convmlp} and Q-Former~\cite{li2023blip}.

        \noindent\textbf{Rescale-MLP}: With the encoder output sequence $\mathbf{h} \in \mathbb{R}^{\tau \times d}$, we first reshape the embedding to match the desired sequence length for the LLM embedding layers. Specifically, the reshaped embedding is given by $\mathbf{h'} \in \mathbb{R}^{(\tau/s) \times (d \cdot s)}$, where $s$ is the scaling factor that adjusts the original sequence length $\tau$ to $\tau/s$.

        Subsequently, we apply three MLP layers sequentially to transform the embedding dimensions. These layers map the embedding size from $d \cdot s$ to $d$, then from $d$ to $4d$, and finally from $4d$ to $\gamma$, where $\gamma$ denotes the embedding size of the subsequent LLM. We apply two SiLU activations~\cite{elfwing2018sigmoid} between the MLP layers. In our experiments, we evaluated different scaling factors $s$, which resulted in varying sequence lengths derived from speech features for integration with LLMs, where a larger $s$ produces fewer audio tokens.

        \noindent\textbf{Conv-1D}: We employ a 1-dimensional convolutional layer followed by a single MLP layer for feature reorganization. Specifically, we use a kernel size of 8 with a stride of 8. The number of output channels for the 1-D convolutional layer is set to $\gamma$, which corresponds to the embedding size of LLMs. Consequently, the output sequence length is downsampled to approximately $\tau/8$. Following this, we apply a single MLP layer with both input and output embedding sizes equal to $\gamma$.

        Unlike~\citet{li2023convmlp}, we utilize a 1D convolutional layer instead of 2D, as our focus is on processing 1D audio features. The downsampled output sequence length facilitates faster training and inference when integrated with LLMs.

        \noindent\textbf{Q-Former}: Instead of using the original Q-Former proposed in~\citet{li2023blip}, we adopt the window-based Q-Former as introduced in~\citet{tang2023salmonn}. The window-based Q-Former is designed to improve efficiency, particularly when the encoder sequence length is long, which can result in significant memory consumption due to the computation of cross-attentions within transformer layers. Similarly, we randomly initialize the Q-Former layers, and the Query tokens are trainable parameters integrated into the network structure.

    \subsection{Large Language Model}

        As Singlish incorporates a mix of multiple dialects and involves code-switching in certain data samples, we aim to examine whether a more advanced multilingual large language model can improve spoken understanding tasks in this scenario. To this end, our study compares both Gemma models~\cite{gemma} and Llama models~\cite{dubey2024llama}. Additionally, we also investigate the impact of model size on performance. Specifically, the speech token embeddings are concatenated with the token embeddings from prompts using the following template after the embedding layer of the LLMs: ``<speech> $\mathbf{s}_{tokens}$ </speech> $t_{prompt}$''. Additionally, the corresponding LLM chat template is applied to align with the instruction-tuned capabilities of the LLM.

    \subsection{Training Method}

        In our approach, the encoder and adaptor are fully fine-tuned, while the decoder undergoes LoRA tuning~\cite{hu2021lora}, with the model designed for autoregressive text generation tasks. 
        In our experiments, we explored alternative training setups. We observed that without adapting the encoder, it becomes challenging to learn new accents, phonetics, and local terms, which are prevalent in Singlish but beyond standard English. Additionally, replacing LoRA tuning with full fine-tuning of the LLM significantly impacts its original capabilities, resulting in diminished performance in following diverse instructions. This is why we adopt the above training configuration in all experiments. Furthermore, some existing studies have investigated various alternative approaches to address these challenges through multi-task learning, multi-stage training, and others~\cite{tang2023salmonn,he2024meralion,chu2024qwen2,lu2024developing}.

    \begin{table*}[htb]
        \centering
        \begin{adjustbox}{width=0.98\textwidth,center}
        \begin{tabular}{ l | c  c  c  c  c | c }
        \toprule
         \multirow{1}{*}{\textbf{Dataset}} & \textbf{SingAudioLLM} & \textbf{Qwen2-Audio-Instruct} & \textbf{WavLLM} & \textbf{Qwen-Audio-Chat} & \textbf{SALMONN} & \multirow{1}{*}{\textbf{Cascade Model}} \\ \midrule
         
        \multicolumn{3}{l}{\sethlcolor{color1!100}\hl{\textbf{Automatic Speech Recognition (ASR)}}{$(\downarrow)$}}
        \\  \hdashline
        
         {MNSC-ASR PART 1}  & \textbf{5.1} & 7.2 & 10.1 & 10.6 & 9.3 & 6.9 \\ 
         {MNSC-ASR PART 2}  & \textbf{5.1} & 19.1 & 44.6 & 45.5 & 42.3 & 31.9 \\ 
         {MNSC-ASR PART 3}  & \textbf{28.1} & 35.1 & 75.4 & 64.1 & 65.7 & 30.0 \\ 
         {MNSC-ASR PART 4}  & \textbf{39.9} & 56.1 & 114.4 & 117.3 & 75.9 & 47.5 \\ 
         {MNSC-ASR PART 5}  & \textbf{20.1} & 27.9 & 39.8 & 30.2 & 34.8 & 22.0 \\ 
         {MNSC-ASR PART 6}  & \textbf{14.5} & 22.5 & 42.5 & 31.4 & 24.9 & 17.5 \\ 
         
        \midrule

        \multicolumn{3}{l}{\sethlcolor{color2!100}\hl{\textbf{Spoken Question Answering (SQA)}}{$(\uparrow)$}}
        \\  \hdashline

         {MNSC-SQA PART 3}  & \textbf{50.0} & 42.0 & 45.2 & 32.2 & 40.6 & 49.0 \\ 
         {MNSC-SQA PART 4}  & 51.8 & 39.6 & 46.6 & 37.8 & 36.6 & \textbf{53.8} \\ 
         {MNSC-SQA PART 5}  & \textbf{62.0} & 51.6 & 50.8 & 47.8 & 44.6 & 57.8 \\ 
         {MNSC-SQA PART 6}  & \textbf{64.0} & 53.6 & 62.2 & 51.4 & 46.8 & \textbf{64.0} \\ 
         \midrule

        \multicolumn{3}{l}{\sethlcolor{color3!100}\hl{\textbf{Spoken Dialogue Summarization Answering (SDS)}}{$(\uparrow)$}}
        \\ \hdashline

         {MNSC-SDS PART 3}  & \textbf{46.4} & 33.8 & 31.6 & 16.4 & 9.0 & 37.4 \\ 
         {MNSC-SDS PART 4}  & \textbf{45.4} & 24.8 & 31.6 & 16.0 & 7.0 & 36.0 \\ 
         {MNSC-SDS PART 5}  & \textbf{50.0} & 40.4 & 45.2 & 28.2 & 17.2 & 49.0 \\ 
         {MNSC-SDS PART 6}  & \textbf{61.4} & 46.2 & 49.4 & 40.4 & 24.2 & 57.2 \\ 
         
         \midrule

        \multicolumn{3}{l}{\sethlcolor{color4!100}\hl{\textbf{Paralinguistic Question Answering (PQA)}}{$(\uparrow)$}}
        \\ \hdashline

         {MNSC-Gender-Sentence}  & \textbf{97.1} & 68.6 & 49.0 & 57.5 & 59.8 & 35.4 \\ 
         {MNSC-Gender-Dialogue}  & \textbf{91.8} & 61.5 & 46.8 & 37.1 & 43.0 & 25.6 \\ 
         {MNSC-Accent-Sentence}  & \textbf{95.2} & 2.5 & 2.9 & 3.7 & 2.5 & 12.4 \\ 
         {MNSC-Accent-Dialogue}  & \textbf{74.9} & 0.8 & 0.3 & 0.6 & 0.1 & 10.3 \\ 
         
         \midrule  

        \multicolumn{3}{l}{\textbf{Hold-Out Test Set}}
        \\ \hdashline

         {CN-College-Listen-Test}  & 79.7 & 74.5 & 65.4 & 63.3 & 50.9 & \textbf{85.2} \\
         {Singapore Public Speech QA}  & 56.3 & 58.3 & 58.5 & 63.2 & 59.2 & \textbf{64.9} \\
         SEAME-Dev-SGE & \textbf{38.8} & 54.9 & 122.0 & 105.6 & 101.9 & 54.9 \\
         SEAME-Dev-MAN & \textbf{48.4} & 55.2 & 129.1 & 87.8 & 127.2 & 55.2\\
         \bottomrule
        
        \end{tabular}
        \end{adjustbox}
        \caption{Main results of four AudioLLMs and one cascade model. The word-error-rate (WER) for ASR tasks is the lower the better\textsubscript{$(\downarrow)$}. The best performance is highlighted in bold.}
        \label{tab:results}
    \end{table*}

    \begin{table}[t]
        \centering
        \begin{adjustbox}{width=0.49\textwidth,center}
        \begin{tabular}{ l | c  c | c  }
        \hline
         \multirow{1}{*}{\textbf{Dataset}} & \textbf{SingAudioLLM} & \textbf{FT-Whisper} & \multirow{1}{*}{\textbf{Whisper-Large-v3}} \\ 
         
         \hline
        
         {MNSC-ASR PART 1}  & \textbf{4.4} & \textbf{4.4} & 6.9 \\ 
         {MNSC-ASR PART 2}  & 4.2 & \textbf{3.8} & 31.9 \\ 
         {MNSC-ASR PART 3}  & 23.7 & \textbf{18.8} & 30.0 \\ 
         {MNSC-ASR PART 4}  & 34.8 & \textbf{29.8} & 47.5 \\ 
         {MNSC-ASR PART 5}  & \textbf{19.4} & 20.6 & 22.0 \\ 
         {MNSC-ASR PART 6}  & \textbf{13.4} & 24.0 & 17.5 \\ 
         
         \hdashline

         SEAME-Dev-SGE & 31.7 & \textbf{27.2} & 54.9 \\
         SEAME-Dev-MAN & 35.2 & \textbf{28.8} & 55.2\\
         \hline
        
        \end{tabular}
        \end{adjustbox}
        \caption{Comparison of three models on Singlish ASR performance. The SingAudioLLM and FT-Whisper are purely fine-tuned with ASR data from NMSC.}
        \label{tab:results_asr_only}
    \end{table}

\section{Experiments}

    We train SingAudioLLM on the full MNSC corpus to evaluate its effectiveness for experiments. For the main results, we employ Whisper-large-v3 as the encoder, Conv-1D as the modality adaptor, and Gemma2-9B-Instruct as the decoder, selected for their robust performance and efficiency. Additionally, we conduct an ablation study using alternative variants to provide further insights. 
    For evaluation, in addition to standard in-domain evaluations using MNSC subsets, we include three external evaluation sets: CN-College-Listen-Test~\cite{wang2024audiobench,hu2024wavllm}, Singapore Public Speech QA~\cite{wang2024audiobench}, and SEAME~\cite{lyu2010seame}. These evaluation sets focus on English comprehension, question answering based on Singapore public speeches, and code-switching Singlish ASR for English and Mandarin, respectively, providing a comprehensive understanding of the model's generalizability. All evaluation metrics adhere to the original design, utilizing either Word Error Rate (WER) or model-as-judge scores.

    \noindent\textbf{Baselines}. For comparison, we include four end-to-end multimodal models Qwen2-Audio-Instruct~\cite{chu2024qwen2}, WavLLM~\cite{hu2024wavllm}, Qwen-Audio-Chat~\cite{chu2023qwen} and SALMONN~\cite{tang2023salmonn}, and a cascaded model that integrates ASR outputs from Whisper-Large-v3 into the Llama-3-8B-Instruct model. We also fine-tuned Whisper using our ASR subsets to evaluate its performance with adaptation as a standalone ASR model. 

    \begin{figure*}[t]
         \centering
             \includegraphics[width=0.975\textwidth]{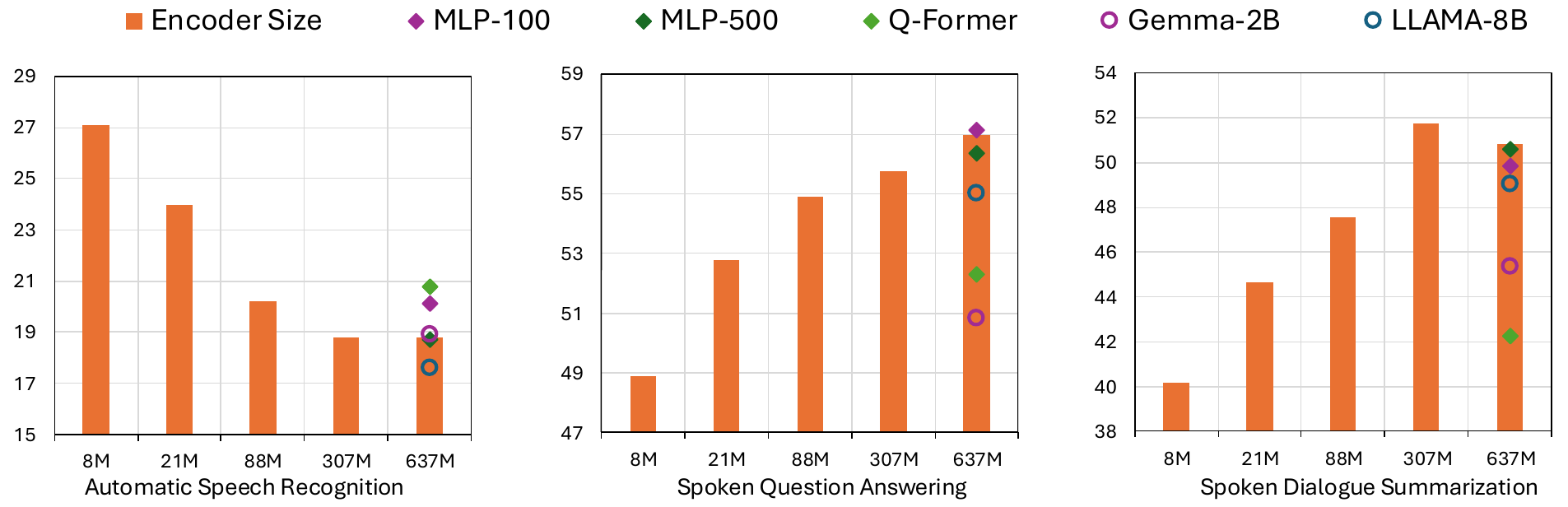}
            \caption{
                We conducted a study to assess the effectiveness of different encoder sizes, adaptor types, decoder sizes, and decoder types across three tasks, reporting the average performance for each task. Encoder sizes were evaluated using five variants ranging from 8M to 637M parameters, represented along the x-axis. The comparison of adaptors and decoders is represented with diamonds and circles.
            }
            \label{fig:3}
    \end{figure*}

    \noindent\textbf{Results and Analysis}. 
    As shown in Table~\ref{tab:results}, Sing-AudioLLM demonstrates superior in-domain performance compared to all other models. For ASR tasks, Whisper provides a strong baseline. Notably, the MNSC-ASR-PART 1 dataset achieves strong results, with all models reaching below 10 WER, as it features read speech without significant use of local terminology. However, other scenarios, such as local adaptation in PART 2 and dialogue-based tasks, require substantial adaptation. In these cases, the LLM used as a decoder proves effective, successfully learning and decoding Singapore-specific terms.

    For SQA and SDS tasks, the cascade model outperforms other AudioLLMs by a clear margin. This highlights the advantage of decoupling transcription generation and instruction following, preserving the LLM’s reasoning capabilities and flexibility. The cascade model handles noisy inputs more effectively, leading to second-best performance. Nevertheless, domain adaptation remains critical, as evidenced by the improved results from our model.
    
    For PQA tasks, gender inference shows moderate success with existing AudioLLMs. Among the baselines, Qwen2-Audio-Instruct delivers the best results but struggles with Singlish speakers, especially in multi-speaker settings. Accent recognition, in particular, requires significant domain adaptation to achieve reasonable performance. All existing AudioLLMs fail to generalize effectively to our scenarios.
    
    In zero-shot evaluations on hold-out datasets, SingAudioLLM demonstrates strong transferability from Singlish datasets to general English domains. For instance, it achieves the highest performance on the CN-College-Listen-Test dataset among all AudioLLMs, despite not being trained on general domain data. Additionally, SEAME datasets show strong zero-shot performance, outperforming Whisper models. These results indicate that MNSC datasets generalize well to other accents and code-switching scenarios.

    \noindent\textbf{ASR Performance}. 
    To further evaluate the effectiveness of the MNSC datasets beyond their use in AudioLLMs, we adapted the Whisper-large model and compared it to our SingAudioLLM, which was trained exclusively on MNSC-ASR datasets without incorporating additional Singlish understanding tasks. As shown in Table~\ref{tab:results_asr_only}, the MNSC datasets significantly enhance Whisper's adaptation to the Singlish domain, demonstrating strong performance even in zero-shot evaluations on SEAME datasets. Additionally, ASR-only AudioLLM models outperform multitask AudioLLMs as task-specific solutions, highlighting the effectiveness of task-specific models in maintaining a sharper focus on the target objective (e.g ASR). 

    \noindent\textbf{Impact of Encoder}. For the decoder size study, we varies the encoder configurations of Whisper while keeping other settings fixed, using Conv-1D as the adaptor and Gemma-9B as the LLM decoder. As shown in Figure~\ref{fig:3}, performance generally degrades across all three tasks with smaller encoders. An exception is observed with Whisper-large and Whisper-medium, where ASR performance remains comparable. Notably, Whisper-medium outperforms Whisper-large in dialogue summarization tasks, indicating that both models can generate robust semantic representations of speech content. However, with smaller encoders, performance drops significantly, with noticeable degradations across tasks. These findings suggest that a strong speech encoder is critical to effectively integrating LLMs.

    \noindent\textbf{Effect on Adaptors}. We evaluated three adaptor variants, as introduced in Section~\ref{sec:adaptor}, with results illustrated by the dotted points in Figure~\ref{fig:3}. MLP-100 ($s=15$) and MLP-500 ($s=3$) demonstrated comparable performance to the Conv-1D model. The Conv-1D model compresses the input sequence length of audio embeddings to approximately 180 tokens, while MLP-100 and MLP-500 compress to 100 and 500 tokens, respectively. Theoretically, longer input sequences to the LLM decoder increase generation time and latency, and may not be necessary to effectively encode audio information. Considering the balance between encoding efficiency, generation speed, and overall task performance, Conv-1D is selected as the default adaptor.
    We found no significant advantage with Q-Former. One possible reason is that we fine-tune the speech feature extractor with full parameters, whereas BLIP-2~\cite{li2023blip} keeps the feature extractor frozen when using Q-Former as an adaptor. This difference in parameter tuning may explain the lack of performance improvement with Q-Former in our setup.

    \noindent\textbf{LLM Decoder Size and Type}. We compared two sizes of LLM decoders: Gemma-9B and Gemma-2B. The results vary across tasks. For ASR, the decoder size has minimal impact on performance. However, for understanding and reasoning tasks, such as question answering and summarization, the larger decoder size significantly outperforms the smaller one. This suggests that ASR primarily relies on superficial features like phoneme and phonetic information, which do not require a large language model for complex reasoning. This also explains why CTC-based models~\cite{kim2017joint} continue to achieve strong performance in certain scenarios compared to LM-based approaches. In contrast, reasoning and summarization tasks benefit substantially from the enhanced capabilities of a larger LLM.
    We also evaluated LLAMA-3-8B-Instruct as the decoder. Consistent with earlier observations, it demonstrated inferior performance on SQA and SDS tasks, indicating weaker reasoning capabilities compared to the Gemma models. However, LLAMA-3-8B outperformed Gemma-2B in the ASR task, the stronger decoder does not have obvious effect on ASR task.

\section{Conclusion}

    This study addresses the critical gap in spoken Singlish research by introducing the largest standardized spoken Singlish corpus, \textbf{MNSC}, and the novel \textbf{SingAudioLLM} model. The proposed datasets and model demonstrate significant advancements in multilingual and multimodal audio understanding, achieving state-of-the-art performance across diverse tasks. By releasing the datasets, model, and code, we aim to foster future research and innovation in Singlish language processing and beyond.

\section*{Limitations}

    The primary limitation of this work arises from the data preparation of the original National Speech Corpus, which was not specifically designed for dialogue tasks, and the variability in recording environments, as discussed in Section~\ref{sec:4}. To ensure high data quality, we prioritized releasing only data with high confidence. However, a substantial portion of the data was excluded from this release due to issues with alignment noise and inappropriate annotations. This highlights that not all data sources were fully utilized, leaving room for improvement by incorporating more available data in the future.

\section*{Acknowledgement}

    This research is supported by the National Research Foundation, Singapore and Infocomm Media Development Authority, Singapore under its National Large Language Models
    Funding Initiative. Any opinions, findings and conclusions or
    recommendations expressed in this material are those of the
    author(s) and do not reflect the views of National Research
    Foundation, Singapore and Infocomm Media Development
    Authority, Singapore.

    We want to thank Ziyi Xu for helping with data preparation on SEAME datasets and ShuangHong Huang for providing useful feedback on the manuscript.

\bibliography{anthology,custom}

\appendix

\section{MNSC Samples}

    The following table presents examples from various MNSC subsets.

\section{Hardward}

    Experiments were conducted on multiple DGX nodes featuring 8 NVIDIA H100 GPUs with 80GB VRAM each. These specifications ensure efficient training and evaluation across various tasks.

\section{License}
    
    In terms of licensing, the MNSC datasets follow the NSC dataset to be under the Singapore Open Data Licence\footnote{\url{https://data.gov.sg/open-data-licence}}, allowing free use with proper attribution. Restrictions include no rights to personal data, unauthorized third-party rights, or intellectual property.

\begin{sidewaystable*}[htb]
\centering
\begin{adjustbox}{width=1.00\textwidth,center}
\begin{tabular}{|p{4.5cm}|p{4cm}|p{6cm}|p{6cm}|p{6cm}|}
\hline
\textbf{Dataset} & \textbf{Instruction} & \textbf{Reference Response} & \textbf{Sing-AudioLLM Response} & \textbf{Qwen2-Audio-Instruct Response} \\ \hline

MNSC-ASR-PART 1 & Transcribe the spoken conversation into written text. & in their hut the fridge is piled with frozen foodstuff dating months back which they refuse to throw away & In their hut, the fridge is piled with frozen foodstuff dating months back which they refused to throw away. & In their hut, the fridge is packed with frozen food dating months back, which they refuse to throw away. \\ \hline

MNSC-ASR-PART 2 & (Paraphrase of above) & there is a food court selling Chai Tow Kway behind Evon's house & There is a food court selling Chai Tow Kway behind Irl's house. & The food court selling chai tow kway is located behind Evon's house. \\ \hline

MNSC-ASR-PART 3,5,6 & (Paraphrase of above) 
& <Speaker1>: also I think because there is a bar nearby right in Singapore \#Sentosa\# there are also many bars in the at the beach so if there are bars right there are bound to be some (um) glasses shards on the sand so I guess it's to prevent you know getting cuts on the leg and the feet right that's why we wear shoes to prevent cuts <Speaker2>: [oh] I see
& <Speaker1>: Also, I think because there is a bar nearby right in Singapore. \#sentosa\# There are also many bars in the at the beach. So if they have bars right, there are bound to be some (um) glasses shards on the sand. so I guess it's to prevent you know getting cuts on the leg. <Speaker2>: Ya. <Speaker1>: On the feet, right? that's why we wear shoes to prevent cuts. <Speaker2>: [oh] I see.
& also, i think because there's a bar nearby right in singapore sentosa, there are also many bars in the at the beach. so if they are bars right, they are bound to be some glasses, sharks, on the sand. so i guess it's to prevent, you know, getting cuts on the leg and feet, right? that's why we wear shoes to prevent cuts.
\\ \hline

MNSC-ASR-PART 4 & (Paraphrase of above) 
& <Speaker1>: and I remember the restaurant is quite nice [one] <Speaker2>: [oh] vivo city \begin{CJK*}{UTF8}{gbsn} 比较好 \end{CJK*} right <Speaker1>: ya \begin{CJK*}{UTF8}{gbsn} 那边有很多 \end{CJK*} nice shops <Speaker2>: !aiya! raffles city \begin{CJK*}{UTF8}{gbsn} 有点远 \end{CJK*} acai \begin{CJK*}{UTF8}{gbsn} 是哪里 \end{CJK*} again <Speaker1>: acai \begin{CJK*}{UTF8}{gbsn} 是 \end{CJK*} holland village <Speaker2>: \begin{CJK*}{UTF8}{gbsn} 哦那个比较靠近我们的家 \end{CJK*} <Speaker1>: yes but Brotzeit \begin{CJK*}{UTF8}{gbsn} 的是 九块 \end{CJK*} an hour so is not bad <Speaker2>: (um) then weekend \begin{CJK*}{UTF8}{gbsn} 有没有多一点钱 \end{CJK*}

& <Speaker1>: And I remember the restaurant is quite nice [one] <Speaker2>: !wah!, Vivo City, \begin{CJK*}{UTF8}{gbsn} 比较好 \end{CJK*} right. <Speaker1>: Ya, \begin{CJK*}{UTF8}{gbsn} 那边有很多 \end{CJK*} nice shops. <Speaker2>: !aiya!, Raffles City \begin{CJK*}{UTF8}{gbsn} 有点远阿拉赛是哪里 \end{CJK*} again? <Speaker1>: Asasay, \begin{CJK*}{UTF8}{gbsn} 是 \end{CJK*} Holland Village. <Speaker2>: [oh] \begin{CJK*}{UTF8}{gbsn} 那个比较近我们的家啊 \end{CJK*} <Speaker1>: Yes, but Broadside \begin{CJK*}{UTF8}{gbsn} 的是九点 \end{CJK*} an hour, so it's not bad. <Speaker2>: (uh) then weekend, \begin{CJK*}{UTF8}{gbsn} 有没有多一点钱 \end{CJK*}.

& and i remember the restaurant is quite nice one vivocity city picture house right yeah nothing you know her door nice shots  yeah reference city ooh din  asians in ashiya suna nee again arsene is holly village zero like a video counting moment that is yeah yes but broadside does it still quiet an hour though so it is not bad then we can you know do it in chen \\

\hline
\end{tabular}
\end{adjustbox}
\caption{Examples from MNSC ASR datasets and responses from models.}
\end{sidewaystable*}

\begin{sidewaystable*}[htb]
\centering
\begin{adjustbox}{width=1.00\textwidth,center}
\begin{tabular}{|p{2.5cm}|p{4cm}|p{5cm}|p{5cm}|p{5cm}|}
\hline
\textbf{Dataset} & \textbf{Instruction} & \textbf{Reference Response} & \textbf{Sing-AudioLLM Response} & \textbf{Qwen2-Audio-Instruct Response} \\ \hline

MNSC-SQS-PART 3-6 
& What does the first speaker share about their experience with Jay Chou and their opinion of him compared to JJ Lin? 
& The first speaker describes being very close to Jay Chou, about three meters away, and mentions that it was one of the best experiences for them as a fan. They think Jay Chou is a very good singer and prefer his songs over JJ Lin's, even though JJ Lin is more popular in Singapore. 
& The first speaker shares that they were very close to Jay Chou, about three meters away, and expresses their admiration for him as a singer, considering him better than JJ Lin. 
& The first speaker shares an interesting story about encountering Jay Chou and his reaction to his height, followed by their opinion that Jay Chou is very good and deserves a chance in Singapore, similar to JJ Lin. \\ \hline

MNSC-SDS-PART 3-6 
& Generate a summary that distills the dialogue into its most important discussions and decisions.
& The speakers discuss a unique drink combination of Baileys and sweet green tea. Speaker1 suggests it as a must-try, while Speaker2 expresses curiosity and interest in giving it a shot.
& Speaker1 and Speaker2 discussed a combination of Baileys, a type of drink, with green tea. Speaker1 mentioned that it was supposedly the best combo despite sounding weird, but wasn't sure if it was specifically Baileys green tea or just any sweet green tea. Speaker2 confirmed that it would be Bailey's with Pokka green tea and asked if it was worth trying, expressing interest in trying Hong-Tao instead.
& The audio's key points revolve around the preferences of drinks, specifically green tea, and the opinions about the best combination or type of green tea.
\\ \hline

\end{tabular}
\end{adjustbox}
\caption{Examples from MNSC SQS, SDS datasets and responses from models.}
\end{sidewaystable*}

\begin{sidewaystable*}[htb]
\centering
\begin{adjustbox}{width=1.00\textwidth,center}
\begin{tabular}{|p{3cm}|p{4cm}|p{4cm}|p{4cm}|p{4cm}|}
\hline
\textbf{Dataset} & \textbf{Instruction} & \textbf{Reference Response} & \textbf{Sing-AudioLLM Response} & \textbf{Qwen2-Audio-Instruct Response} \\ \hline

MNSC-Gender-Sentence 
& How would you describe the speaker's gender in this clip?
& The speaker sounds like a female.
& The speaker is a female.
& The speaker is female.
\\ \hline

MNSC-Gender-Dialogue 
& What are the genders of the two speakers in the dialogue?
& The first speaker is female, and the second speaker is male.
& The first speaker is female and the second speaker is male.
& The first speaker is female, and the second speaker is male.
\\ \hline

MNSC-Accent-Sentence 
& What can you tell about the speaker's accent?
& The speaker communicates fluently in English with a Singaporean Indian accent, marked by distinctive local intonations and the occasional use of Singlish expressions.
& The speaker speaks fluent English with a Singaporean Indian-Speaking accent, characterized by local intonation and occasional Singlish expressions.
& The speaker has an English accent. 
\\ \hline

MNSC-Accent-Dialogue 
& How would you describe the accents of the two speakers?
& The first speaker speaks good English with a Singapore accent, while the second speaker has a Malay accent, likely from the Malay-Speaking community of Singapore.
& The first speaker has a Malay accent, likely from the Malay-Speaking community of Singapore, while the second speaker speaks good English with a Singapore accent.
& The first speaker has an Australian accent and the second speaker has an English accent.
\\ \hline

\end{tabular}
\end{adjustbox}
\caption{Examples from MNSC PQA datasets and responses from models.}
\end{sidewaystable*}
    
\end{document}